\documentclass[a4paper,12pt]{article}
\usepackage{amsmath}
\usepackage{algorithmic}
\usepackage{algorithm}
\usepackage{graphicx,graphics}
\usepackage{hyperref}
\title{Feature-Based Matrix Factorization}
\author{
  Tianqi Chen, Zhao Zheng, Qiuxia Lu, Weinan Zhang, Yong Yu\\
  \small{\{tqchen,zhengzhao,luqiuxia,wnzhang,yyu\}@apex.sjtu.edu.cn}\\
  \small{Apex Data \& Knowledge Management Lab}\\
  \small{Shanghai Jiao Tong University}\\
  \small{800 Dongchuan Road, Shanghai 200240 China}\\
  \small{Project page: http://apex.sjtu.edu.cn/apex\_wiki/svdfeature}
}
\date{2011-07-11\small{(version 1.1)}}
\begin{document}

\maketitle
\abstract{
Recommender system has been more and more popular and widely used in many applications recently. The increasing information available, not only in quantities but also in types, leads to a big challenge for recommender system that how to leverage these rich information to get a better performance. Most traditional approaches try to design a specific model for each scenario, which demands great efforts in developing and modifying models. In this technical report, we describe our implementation of feature-based matrix factorization. This model is an abstract of many variants of matrix factorization models, and new types of information can be utilized by simply defining new features, without modifying any lines of code. Using the toolkit, we built the best single model reported on track 1 of KDDCup'11.
}

\section{Introduction}

Recommender systems that recommends items based on users interest has become
more and more popular among many web sites. Collaborative Filtering(CF) techniques
that behind the recommender system have been developed for many years and
keep to be a hot area in both academic and industry aspects. Currently CF problems face
two kinds of major challenges: how to handle large-scale dataset and how to
leverage the rich information of data collected.

Traditional approaches to solve these problems is to design specific models
for each problem, i.e writing code for each model, which demands great
efforts in engineering. Matrix factorization(MF) technique is one of the most popular
method of CF model, and extensive study has been made in different variants of matrix
factorization model, such as \cite{koren:knn}\cite{koren:temporal} and \cite{paterek2007}.
However, we find that the majority of matrix factorization models
share common patterns, which motivates us to put them together into one.
We call this model feature-based matrix factorization. Moreover, we write a toolkit
for solving the general feature-based matrix factorization problem, saving the efforts of engineering for detailed kinds of model.
Using the toolkit, we get the best single model on track 1 of KDDCup'11\cite{kddcupdata}.

This article serves as a technical report for our toolkit of feature-based
matrix factorization\footnote{http://apex.sjtu.edu.cn/apex\_wiki/svdfeature}.
We try to elaborate three problems in this report, i.e, what the model is, how can we
use such kind of model, and additional discussion of issues in engineering 
and efficient computation.

\section{What is feature based MF}
In this section, we will describe the model of feature based matrix factorization, starting from the example of linear regression, and then going to
the full definition of our model.

\subsection{Start from linear regression}\label{sec:linear}
Let's start from the basic collaborative filtering models. The very
baseline of collaborative filtering model may be the baseline models just considering
the mean effect of user and item. See the following two models.
\begin{equation}\label{eq:lr1}
  \hat{r}_{ui} = \mu + b_u
\end{equation}
\begin{equation}\label{eq:lr2}
  \hat{r}_{ui} = \mu + b_u + b_i
\end{equation}
Here $\mu$ is a constant indicating the global mean value of rating.
Equation \ref{eq:lr1} describe a model considering users' mean effect while
Equation \ref{eq:lr2} denotes items' mean effect. A more complex model considering
the neighborhood information\cite{koren:knn} is as follows
\begin{equation}\label{eq:lr3}
  \hat{r}_{ui} = \mu + b_i + b_u + |R(u)|^{-\frac{1}{2}}\sum_{j\in R(u)} s_{ij}(r_{uj} - \bar{b}_u)
\end{equation}
Here $R(u)$ is the set of items user $u$ rate, $\bar{b}_u$ is a user average rating pre-calculated. $s_{ij}$ means the similarity parameter from $i$ to $j$. $s_{ij}$ is a parameter that we train from data instead of direct calculation using memory based methods.
Note $\bar{b}_u$ is different from $b_u$ since it's pre-calculated. This is a neighborhood model
that takes the neighborhood effect of items into consideration.

Assuming we want to implement all three models, it seems to be wasting to write code for each
of the model. If we compare those models, it is obvious that all the three models are special cases of
linear regression problem described by Equation \ref{eq:lr}
\begin{equation}\label{eq:lr}
  y = \sum_{i} w_i x_i
\end{equation}
Suppose we have $n$ users, $m$ items, and $h$ total number of possible $s_{ij}$ in equation \ref{eq:lr3}.
We can define the feature vector $x = [x_0,x_1,\cdots,x_{n+m+h}]$ for user item pair $<u,i>$ as follows
\begin{equation}
  x_k = \left\{
    \begin{array}{ll}
      Indicator( u == k ) & k < n\\
      Indicator( i == k - n ) & n \leq k < n+m \\
      0  &  k \geq m+n, j \notin R(u),\mbox{$s_{ij}$ means $w_k$} \\
      |R(u)|^{-\frac{1}{2}}(r_{uj} - \bar{b}_u)  &  k \geq m+n, j\in R(u),\mbox{$s_{ij}$ means $w_k$} \\
    \end{array}
    \right.
\end{equation}
The corresponding layout for weight $w$ shown in equation \ref{eq:lr_layout}. Note that choice of pairs $s_{ij}$ can be flexible.
We can choose only possible neighbors instead of enumerating all the pairs.
\begin{equation}\label{eq:lr_layout}
  w = [ b_u(0), b_u(1),\cdots, b_u(n), b_i(1), \cdots b_i(m) \cdots s_{ij} \cdots ]
\end{equation}
In other words, equation \ref{eq:lr3} can be reformed as the following form
\begin{equation}\label{eq:lr4}
  \hat{r}_{ui} = \mu + b_i 1 + b_u 1 + \sum_{j\in R(u)} s_{ij} \left[|R(u)|^{-\frac{1}{2}}(r_{uj} - \bar{b}_u)\right]
\end{equation}
where $b_i$, $b_u$, $s_{ij}$ corresponds to weight of linear regression, and the coefficients on the right
of the weight are the input features. In summary, under this framework, the only thing that we need to do
is to layout the parameters into a feature vector. In our case, we arrange first
$n$ features to $b_u$ then $b_i$ and $s_{ij}$, then transform the input data into the format of linear regression
input. Finally we use a linear regression solver to work the problem out.

\subsection{Feature based matrix factorization}
The previous section shows that some baseline CF algorithms are linear regression problem. In this section, we will discuss feature-based generalization for matrix factorization. A basic matrix factorization model is stated in Equation \ref{eq:bmf}:
\begin{equation}\label{eq:bmf}
  \hat{r}_{ui} = \mu +b_u +b_i + p^T_{u} q_{i}
\end{equation}
The bias terms have the same meaning as previous section. We also get two factor term $p_u$ and $q_i$. $p_u$ models the latent peference of user $u$. $q_i$ models the latent property of item $i$.

\begin{figure}
  \centering
  \includegraphics[width=.8\textwidth]{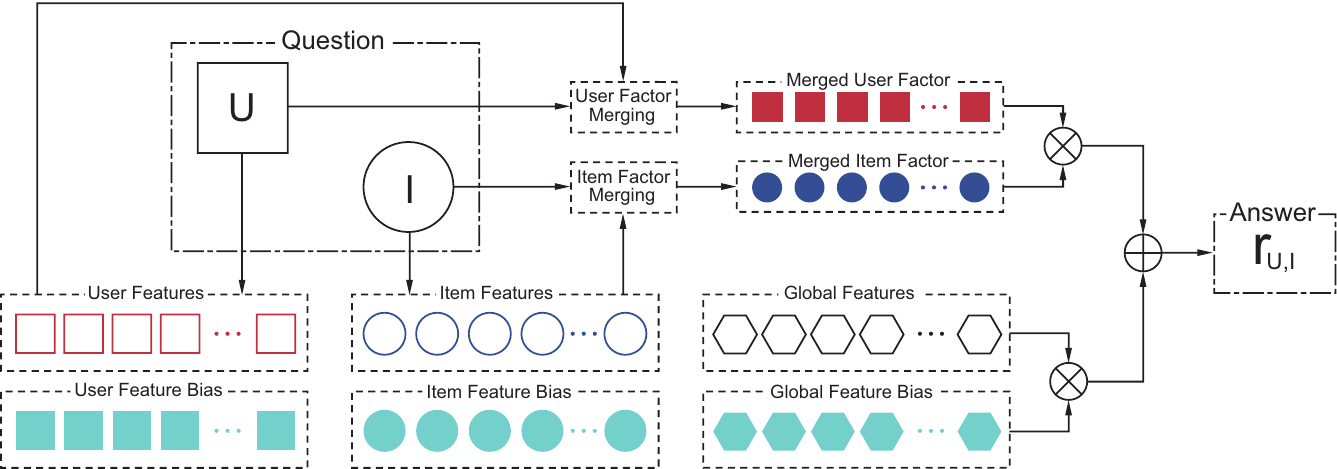}
  \caption{Feature-based matrix factorization}\label{fig:fmf}
\end{figure}

Inspired by the idea of previous section, we can get a direct generalization
for matrix factorization version.
\begin{equation}\label{eq:fmf1}
  \hat{r}_{ui} = \mu + \sum_{j} w_j x_j +b_u +b_i + p^T_{u} q_{i}
\end{equation}
Equation \ref{eq:fmf1} adds a linear regression term to the traditional matrix factorization model.
This allows us to add more bias information, such as neighborhood information and time bias
information, etc. However, we may also need a more flexible factor part.
For example, we may want a time dependent user factor $p_u(t)$ or hierarchical dependent item factor
$q_i(h)$. As we can find from previous section, a direct way to include such flexibility is to use
features in factor as well. So we adjust our feature based matrix factorization as follows
\begin{equation}\label{eq:fmf}
  y =  \mu + \left(\sum_{j} b^{(g)}_j \gamma_j + \sum_{j} b^{(u)}_j \alpha_j + \sum_{j} b^{(i)}_j \beta_j \right)
   +\left( \sum_{j} p_{j} \alpha_j\right)^T \left( \sum_j q_{j} \beta_j\right)
\end{equation}
The input consists of three kinds of features $<\alpha, \beta, \gamma>$,
we call $\alpha$ user feature, $\beta$ item feature and $\gamma$ global feature.
The first part of Equation \ref{eq:fmf}.
The name of these features explains their meanings. $\alpha$ describes the user aspects,
$\beta$ describes the item aspects, while $\gamma$ describes some global bias effect.
Figure \ref{fig:fmf} shows the idea of the procedure.

We can find basic matrix factorization is a special case of Equation \ref{eq:fmf}. For predicting user item pair $<u,i>$,
define
\begin{equation}
  \gamma = \emptyset,\
  \alpha_k = \left\{
    \begin{array}{ll}
      1 & k = u\\
      0 & k \neq u \\
    \end{array}\right.,\
  \beta_k = \left\{
    \begin{array}{ll}
      1 & k = i\\
      0 & k \neq i \\
    \end{array}\right.
\end{equation}

We are not limited to the simple matrix factorization. It enables us to incorporate the neighborhood information to $\gamma$,
and time dependent user factor by modifying $\alpha$. Section
\ref{sec:example} will present a detailed description of this.

\subsection{Active function and loss function}
There, you need to choose an active function $f(\cdot)$ to the output of the feature based matrix factorization. Similarly,
you can also try various of loss functions for loss estimation. The final version of the model is
\begin{equation}
    \hat{r}=  f(y)
\end{equation}
\begin{equation}
  Loss = L( \hat{r}, r ) + regularization
\end{equation}
Common choice of active functions and loss are listed as follows:
\begin{itemize}
\item identity function, L2 loss, original matrix factorization.
  \begin{equation}
    \hat{r}=  f(y) = y
  \end{equation}
    \begin{equation}
      Loss = ( r - \hat{r} )^2 + regularization
  \end{equation}
\item sigmoid  function, log likelihood, logistic regression version of matrix factorization.
  \begin{equation}
    \hat{r} = f(y) =  \frac{1}{1+e^{-y}}
  \end{equation}
    \begin{equation}
      Loss = r \ln \hat{r} + (1-r)\ln (1-\hat{r}) + regularization
  \end{equation}
\item identity  function, smoothed hinge loss\cite{rennie:sgd_mmmf},
  maximum margin matrix factorization\cite{srebro:mmmf}\cite{rennie:sgd_mmmf}.
  Binary classification problem, $r \in \{0,1\}$
  \begin{equation}
      Loss = h\left( (2r-1) y \right) + regularization
  \end{equation}
  \begin{equation}
    h( z ) =
    \left\{
    \begin{array}{ll}
      \frac{1}{2} - z & z \leq 0\\
      \frac{1}{2}(1 - z)^2 & 0<z < 1\\
      0 & z \geq 1 \\
    \end{array}\right.
  \end{equation}
\end{itemize}

\subsection{Model Learning}
To update the model, we use the following update rule
\begin{align}
  p_i &= p_i + \eta \left( \hat{e} \alpha_i \left(\sum_j q_j \beta_j \right) - \lambda_1 p_i \right)\\
  q_i &= q_i + \eta \left( \hat{e} \beta_i  \left(\sum_j p_j \alpha_j\right) - \lambda_2 q_i \right)\\
  b^{(g)}_i&= b^{(g)}_i + \eta \left(\hat{e} \gamma_i - \lambda_3 b^{(g)}_i\right)\\
  b^{(u)}_i&= b^{(u)}_i + \eta \left(\hat{e} \alpha_i - \lambda_4 b^{(u)}_i\right)\\
  b^{(i)}_i&= b^{(i)}_i + \eta \left(\hat{e} \beta_i  - \lambda_5 b^{(i)}_i\right)
\end{align}
Here $\hat{e} = r - \hat{r}$ the difference between true rate and predicted rate. This rule is valid for both logistic
likelihood loss and L2 loss. For other loss, we shall modify $\hat{e}$ to be corresponding gradient. $\eta$ is the
learning rate and the $\lambda$s are regularization parameters that defines the strength of regularization.

\section{What information can be included}\label{sec:example}
In this section, we will present some examples to illustrate the usage of our feature-based
matrix factorization model.
\subsection{Basic matrix factorization}
Basic matrix factorization model is defined by following equation
\begin{equation}
  y = \mu + b_u + b_i + p^T_u q_i
\end{equation}
And the corresponding feature representation is
\begin{equation}
  \gamma = \emptyset,\
  \alpha_k = \left\{
    \begin{array}{ll}
      1 & k = u\\
      0 & k \neq u \\
    \end{array}\right.,\
  \beta_k = \left\{
    \begin{array}{ll}
      1 & k = i\\
      0 & k \neq i \\
    \end{array}\right.
\end{equation}
\subsection{Pairwise rank model}
For the ranking model, we are interested in the order of two items $i,j$ given
a user $u$. A pairwise ranking model is described as follows
\begin{equation}
  P( r_{ui} > r_{uj} ) = sigmoid\left( \mu + b_i - b_j + p^T_u ( q_i - q_j ) \right)
\end{equation}
The corresponding features representation are like this
\begin{equation}
  \gamma = \emptyset,\
  \alpha_k = \left\{
    \begin{array}{ll}
      1 & k = u\\
      0 & k \neq u \\
    \end{array}\right.,\
  \beta_k = \left\{
    \begin{array}{ll}
       1 & k = i\\
      -1 & k = j\\
      0  & k \neq i, k \neq j \\
    \end{array}\right.
\end{equation}
by using sigmoid and log-likelihood as loss function. Note that the feature representation
gives one extra $b_u$ which is not desirable. We can removed it by give high regularization
to $b_u$ that penalize it to $0$.
\subsection{Temporal Information}
A model that include temporal information\cite{koren:temporal} can be described as
follows
\begin{equation}
  y = \mu + b_u(t)+ b_i(t) + b_u + b_i + \left(p_u + p_u(t)\right)^T q_i
\end{equation}
We can include $b_i(t)$ using global feature, and $b_u(t)$, $p_u(t)$ using user feature.
For example, we can define a time interpolation model as follows
\begin{equation}
  y = \mu +b_i+ b^{s}_u \frac{e-t}{e-s} + b^{e}_u \frac{t-s}{e-s} +
\left(p^{s}_u \frac{e-t}{e-s} +
p^{e}_u \frac{t-s}{e-s}\right)^T q_i
\end{equation}
Here $e$ and $s$ mean start and end of the time of all the ratings. A rating that's rated
later will be affected more by $p^e$ and $b^e$ and earlier ratings will be more affected by
$p^s$ and $b^s$. For this model, we can define
\begin{equation}
  \gamma = \emptyset,\
  \alpha_k = \left\{
    \begin{array}{ll}
      \frac{e-t}{e-s} & k = u\\
      \frac{t-s}{e-s} & k = u + n\\
      0 & \mbox{otherwise} \\
    \end{array}\right.,\
  \beta_k = \left\{
    \begin{array}{ll}
       1 & k = i\\
      0  & k \neq i\\
    \end{array}\right.
\end{equation}
Note we first arrange the $p^s$ in the first $n$ features then $p^e$ in next $n$ features.
\subsection{Neighborhood information}
A model that include neighborhood information\cite{koren:knn}
can be described as below:
\begin{equation}
  y = \mu + \sum_{j\in R(u)} s_{ij} \left[|R(u)|^{-\frac{1}{2}}(r_{uj} - \bar{b}_u)\right] + b_u+b_i+ p^T_u q_i
\end{equation}
We only need to implement neighborhood information to global features as described by Section \ref{sec:linear}.
\subsection{Hierarchical information}
In Yahoo! Music Dataset\cite{kddcupdata}, some tracks belongs to same artist. We can include such
hierarchical information by adding it to item feature. The model is described as follows
\begin{equation}
  y = \mu + b_u + b_t + b_a + p_u^T (q_t+q_a)
\end{equation}
Here $t$ means track and $a$ denotes corresponding artist. This model can be formalized as feature-based
matrix factorization by redefining item feature.

\section{Efficient training for SVD++}
Feature-based matrix factorization can naturally incorporate implicit and explicit information.
We can simply add these information to user feature $\alpha$. The model configuration is shown as follows:
\begin{equation}\label{eq:svdppfmf}
  y = bias + \left( \sum_j \xi_j p_j + \sum_j \alpha_j d_j \right)^T\left(\sum_j \beta_j q_j\right)
\end{equation}
Here we omit the detail of bias term. The implicit and explicit feedback information is given by
$\sum_j \alpha_j d_j$, where $\alpha$ is the feature vector of feedback information,
$\alpha_j=\frac{1}{\sqrt{|R(u)|}}$ for implicit feedback, and $\alpha_j=\frac{r_{u,j}- b_u}{\sqrt{|R(u)|}}$
for explicit feedback. $d_j$ is the parameter of implicit and explicit feedback factor. We explicitly state out the implicit and explicit information
in Equation \ref{eq:svdppfmf}.

Although Equation \ref{eq:svdppfmf} shows that we can easily incorporate implicit and explicit information into the model, it's actually
very costly to run the stochastic gradient training, since the update cost is linear to the size of nonzero entries of $\alpha$,
and $\alpha$ can be very large if a user has rated many items. This will greatly slow down the training speed.
We need to use an optimized method to do training. To show the idea of the optimized method, let's first define a derived user implicit and explicit
factor $p^{im}$ as follows:
\begin{equation}
  p^{im} = \sum_{j} \alpha_j d_j
\end{equation}
The update of $d_j$ after one step is given by the following equation
\begin{equation}
  \Delta d_j = \eta \hat{e}\alpha_j \left(\sum_j \beta_j q_j\right)
\end{equation}
The resulted difference in $p^{im}$ is given by
\begin{equation}\label{eq:svdppup}
  \Delta p^{im} = \eta \hat{e} \left(\sum_j\alpha^2_j\right) \left(\sum_j \beta_j q_j\right)
\end{equation}
Given a group of samples with the \emph{same user}, we need to do gradient descent on each of the
training sample. The simplest way is to do the following steps for each sample:
(1) calculate $p^{im}$ to get prediction (2) update all $d_j$ associates with implicit and explicit feedback.
Every time $p^{im}$ has to be recalculated using updated $d_j$ in this way.
However, we can find that to get new $p^{im}$, we don't need to update each $d_j$. Instead, we only need to
update $p^{im}$ using Equation \ref{eq:svdppup}. What's more, we can find there is a relation between $\Delta p^{im}$
and $\Delta d_j$ as follows:
\begin{equation}\label{eq:rel}
  \Delta d_j = \frac{\alpha_j}{\sum_k\alpha_k^2}\Delta p^{im}
\end{equation}
We shall emphasize that Equation \ref{eq:rel} is true even for \emph{multiple updates}, given the condition that
the \emph{user is same} in all the samples.
We shall mention that the above analysis doesn't consider the regularization term.
If L2 regularization of $d_j$ is used during the update as follows:
\begin{equation}
  \Delta d_j = \eta \left(\hat{e}\alpha_j \left(\sum_j \beta_j q_j\right) -\lambda d_j \right)
\end{equation}
The corresponding changes in $p^{im}$ also looks very similar
\begin{equation}
  \Delta p^{im} = \eta \left(\hat{e} \left(\sum_j\alpha^2_j\right) \left(\sum_j \beta_j q_j\right)- \lambda p^{im}\right)
\end{equation}
However, the relation in Equation \ref{eq:rel} no longer holds strictly. But we can still
use the relation since it approximately holds when regularization term is small.
Using the results we obtained, we can develop a fast algorithm for
feature-based matrix factorization with implicit and explicit feedback information. The algorithm is shown in Algorithm \ref{alg:svdpp}.
\begin{algorithm}
  \caption{Efficient Training for Implicit and Explicit Feedback}\label{alg:svdpp}
  \begin{algorithmic}
    \FORALL{user $u$}
    \STATE $p^{im} \gets \sum_j \alpha_j d_j$ \COMMENT{calculating implicit feedback}
    \STATE $p^{old} \gets p^{im}$
    \FORALL{training samples of user $u$}
    \STATE update other parameters, using $p^{im}$ to replace $\sum_j\alpha_j d_j$
    \STATE update $p^{im}$ directly , do not update $d_j$.
    \ENDFOR
    \FORALL{$ i , \alpha_i\neq 0$}
    \STATE $d_i\gets d_i + \frac{\alpha_i}{\sum_k\alpha_k^2}(p^{im} - p^{old})$ \COMMENT{add all the changes back to $d$}
    \ENDFOR
    \ENDFOR
  \end{algorithmic}
\end{algorithm}

We find that the basic idea is to group the data of the same user together, for the same user shares the same implicit and explicit feedback information.
Algorithm \ref{alg:svdpp} allows us to calculate implicit feedback factor only once for a user, greatly saving the computation time.

\section{How large-scale data is handled}
Recommender system confronts the problem of large-scale data in practice. This is
a must when dealing with real problems. For example Yahoo! Music Dataset\cite{kddcupdata}
consists of more than 200M ratings. A toolkit that's robust to input data size is desirable for real applications.

\subsection{Input data buffering}
The input training data is extremely large in real application, we don't try to load all
the training data into memory. Instead, we buffer all the training data through binary format
into the hard-disk. We use stochastic gradient descend to train our model, that is
we only need to linearly iterate over the data if we shuffle our data before buffering.

Therefore, our solution requires the input feature to be previously shuffled, then a buffering program
will create a binary buffer from the input feature. The training procedure reads the data from hard-disk
and uses stochastic gradient descend to train the model. This buffering approach makes the memory
cost invariant to the input data size, and allows us to train models over large-scale of input data so long
as the parameters fit into memory.

\subsection{Execution pipeline}
\begin{figure}
  \centering
  \includegraphics[width=.8\textwidth]{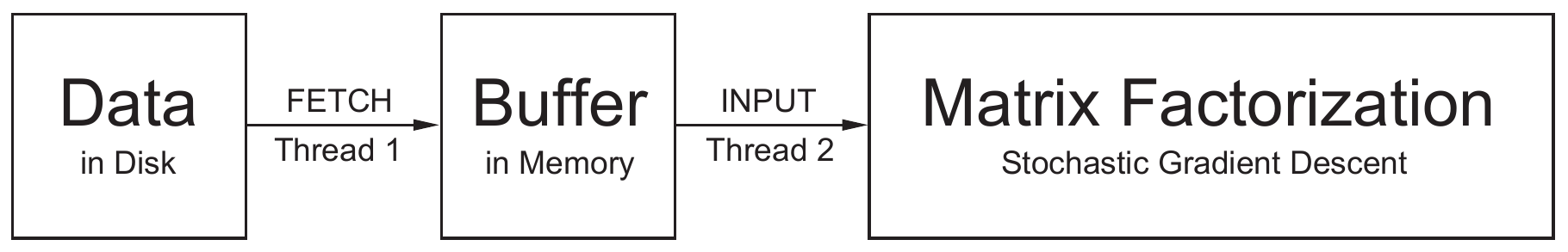}
  \caption{Execution pipeline}\label{fig:pipeline}
\end{figure}

Although input data buffering can solve the problem of large-scale data, it still suffers
from the cost of reading the data from hard-disk. To minimize the cost of I/O, we use a
pre-fetching strategy. We create a independent thread to fetch the buffer data into a memory queue,
then the training program reads the data from memory queue and do training. The procedure is shown in Figure \ref{fig:pipeline}

This pipeline style of execution removes the burden of I/O from the training thread. So long as
I/O speed is similar or faster to training speed, the cost of I/O is negligible, and our our experience
on KDDCup'11 proves the success of this strategy. With input buffering and pipeline execution,
we can train a model with test RMSE=22.16 for
track1 in KDDCup'11\footnote{kddcup.yahoo.com} using less than 2G of memory, without significantly increasing of training time.

\section{Related work and discussion}
The most related work of feature based matrix factorization is
Factorization Machine \cite{rendle:icdm10}. The reader can refer to
libFM\footnote{http://www.libfm.org} for a toolkit for factorization machine.
Strictly speaking, our toolkit implement a \emph{restricted} case of factorization machine
and is more useful in some aspects. We can support global feature that doesn't need to be take into factorization part, which is important for bias features such as user day bias,
neighborhood based features, etc. The divide of features also gives hints for model design.
For global features, we shall consider what aspect may influence the overall rating. For user and item features,
we shall consider how to describe the user preference and item property better.
Our model is also related to \cite{agarwal:kdd} and \cite{stern:www}, the difference is that in feature-based matrix factorization, the user/item feature can associate with temporal information and other context information to better describe the preference or property in current context.
Our current model also has shortcomings.
The model doesn't support multiple distinct factorizations at present. For example, sometimes we may
want to introduce user vs time tensor factorization together with user vs item factorization.
We will try our best to overcome these drawbacks in the future works.
\bibliographystyle{plain}
\bibliography{svd-feature}
\end{document}